\pdfoutput=1

\documentclass[11pt]{article}

\usepackage[]{acl}

\usepackage{times}
\usepackage{latexsym}

\usepackage{dcolumn}
\newcolumntype{L}{D{.}{.}{2,3}}

\usepackage[T1]{fontenc}

\usepackage[utf8]{inputenc}

\usepackage{microtype}

\usepackage{array}
\usepackage[multiple]{footmisc}

\usepackage{fnpct}
\usepackage{caption}
\usepackage{subcaption}

\usepackage{graphicx}
\usepackage{booktabs}
\usepackage{tabularx}
\usepackage{xcolor}
\usepackage{soul}
\usepackage{float}
\usepackage{placeins}
\restylefloat{table}
\usepackage{comment}
\usepackage{rotating}
\usepackage{multirow}
\usepackage{amsmath}
\usepackage{todonotes}
\usepackage{color}
\usepackage{colortbl}

\definecolor{lightblue}{rgb}{.90,.95,1}
\definecolor{light-gray}{gray}{0.9}

\usepackage{lipsum}
\newcommand\blfootnote[1]{%
  \begingroup
  \renewcommand\thefootnote{}\footnote{#1}%
  \addtocounter{footnote}{-1}%
  \endgroup
}

\title{Political Ideology and Polarization:
A Multi-dimensional Approach}

\author{Barea Sinno$^{1*}$\ \ \ \ 
Bernardo Oviedo$^{2*}$\ \ \ \ 
Katherine Atwell$^{3*}$\ \ \ \ \\
\textbf{Malihe Alikhani}$^3$\ \ \ \ 
\textbf{Junyi Jessy Li}$^4$\\
$^1$ Political Science, Rutgers University\\
$^2$ Computer Science, $^4$ Linguistics, The University of Texas at Austin\\
$^3$ Computer Science, University of Pittsburgh\\
{\small {\tt barea.sinno@gmail.com,
bernyoviedo@utexas.edu, kaa139@pitt.edu}}\\
{\small {\tt malihe@pitt.edu, jessy@utexas.edu}}
}

\date{}

\begin{document}
\maketitle\blfootnote{* Equal contribution ordered by first name.}
\begin{abstract}
Analyzing ideology and polarization is of critical importance in advancing our grasp of modern politics. Recent research has made great strides towards understanding the ideological \emph{bias} (i.e., stance) of news media along the left-right spectrum. In this work, we instead take a novel and more nuanced approach for the study of ideology based on its left or right positions on the issue being discussed. Aligned with the theoretical accounts in political science, we treat ideology as a multi-dimensional construct, and introduce the first diachronic dataset of news articles whose ideological positions are annotated by trained political scientists and linguists at the paragraph level. We showcase that, by controlling for the author's stance, our method allows for the quantitative and temporal measurement and analysis of polarization as a multidimensional ideological distance. We further present baseline models for ideology prediction, outlining a challenging task distinct from stance detection. 
\end{abstract}

\section{Introduction}

Political ideology rests on a set of beliefs about the proper order of a society and ways to achieve this order \cite{jost2009political, adorno2019authoritarian, campbell1980american}. In Western politics, these worldviews translate into a multi-dimensional construct that includes: equal opportunity as opposed to economic individualism; general respect for tradition, hierarchy and stability as opposed to advocating for social change; and a belief in the un/fairness and in/efficiency of markets~\cite{jost2009political}.

The divergence in ideology, i.e., polarization, is the undercurrent of propaganda and misinformation~\cite{vicario2019polarization,bessi2016homophily,stanley2015propaganda}. 
It can congest essential democratic functions with an increase in the divergence of political ideologies.
Defined as a growing ideological distance between groups, polarization has waxed and waned since the advent of the American Republic \cite{pierson2020madison}.\footnote{We distinguish ourselves from work that considers other types of polarization, e.g., as a measure of emotional distance~\cite{iyengar2019origins} or distance between political parties~\cite{lauka2018mass}.} 
Two eras---post-1896 and -1990s---have witnessed deleterious degrees of polarization \cite{jenkins2004constituency, jensen2012political}. More recently, COVID-19, the murder of George Floyd, and the Capitol riots have exposed ideological divergences in opinion in the US through  news media and social media.  
With the hope of advancing our grasp of modern politics, we study ideology and polarization
through the lens of computational linguistics by presenting a carefully annotated corpus and examining the efficacy of a set of computational and statistical analyses.

\begin{table}[t!]
\centering
\footnotesize
\begin{tabularx}{\linewidth}{>{\hsize=.5\hsize\linewidth=\hsize} X  |       >{\hsize=1.5\hsize\linewidth=\hsize}X}
\toprule 
Two \vfill dimensions: trade and \vfill economic \vfill liberalism & \textcolor{blue}{\sethlcolor{lightblue}\hl{The U.S. aim is to create a monetary system with enough flexibility to prevent bargain-hungry money from rolling around the world like loose ballast on a ship  disrupting normal trade and currency flows. Nixon goals: dollar, trade stability. This must be accompanied, Washington says, by reduction of [trade] barriers ...}}\\
\rowcolor{light-gray}
One \vfill dimension:  trade \vfill protectionism &  \sethlcolor{lightblue}\hl{The controls program, which Mr. Nixon inaugurated Aug. 15, 1971, has helped to reduce inflation to about 3 percent yearly, and to boost annual U.S. economic growth to more than 7 percent...}\\
\bottomrule
\end{tabularx}
\caption{ Excerpts from news article \#730567 in COHA~\cite{davies2012expanding}.  The first paragraph advocates for liberalism and the reduction of trade barriers. It also has a domestic economic dimension.  The second paragraph, on the contrary, advocates for protectionism and a domestic controls program.} 
\label{tab:ideo_example4}
\end{table}

In contrast to studying the bias or the stance of the author of the text via linguistic framing \cite{kulkarni2018multi,kiesel2019semeval,baly2019multi,baly2020we,chen2020analyzing,stefanov2020predicting}, we study the little explored angle that is nonetheless critical in political science research: \textbf{ideology of the issue (e.g., policy or concept) under discussion}. That is, in lieu of examining the author's stance, we focus on addressing the at-issue content of the text and the ideology that it represents in the implicit social context. 
The nuanced co-existence of stance and  ideology can be illustrated in the following excerpt:
\begin{quote}
\small
\emph{``Republicans and Joe
Biden are making a huge mistake by focusing on cost.  The implication is that government-run health care would be a good thing--a wonderful thing!-- if only we could afford it." \emph{(The Federalist, 9/27/2019)}}
\end{quote}
The author is attacking a liberal social and economic policy; therefore, the ideology being discussed is liberal on two dimensions---social and economic, while the author's stance is conservative. 
Moreover, our novel approach acknowledges that ideology can also \emph{vary} within one article. In Table~\ref{tab:ideo_example4}, we show an example in which one part of an article advocates for trade liberalism, while another advocates for protectionism. 

Together, author stance and ideology inform us not only that there is bias in the media, but also which beliefs are being supported and/or attacked. A full analysis of \emph{polarization} (that reflects a growing distance of political ideology over time) can then be derived if \emph{diachronic} data for both author stance and ideology were available. However, while there has been data for the former (with articles from recent years only)~\cite{kiesel2019semeval}, 
to date, there has been no temporal data on the latter.

In this paper, we present a multi-dimensional framework, and an annotated, diachronic, stance-neutral corpus, for the analysis of ideology in text. This allows us to study polarization as a  \emph{state} of ideological groups with divergent positions on a political issue as well as polarization as a \emph{process} whose magnitude grows over time \cite{dimaggio1996have}. We use proclaimed center, center-left and center-right media outlets who claim to be objective in order to focus exclusively and more objectively on the ideology of the issue being discussed, without the subjectivity of author stance annotation.
We study ideology within every paragraph\footnote{We use automatically segmented paragraphs since the raw texts were not paragraph-segmented.} of an article and aim to answer the following question: which ideological dimension is present and to which 
ideological position does 
it correspond to on the liberal-conservative spectrum. 

Our extensive annotation manual is developed by a political scientist, and the data then annotated by three linguists after an elaborate training phase (Section~\ref{sec:datacollection}). After 150 hours of annotation, we present a dataset of 721 \emph{fully adjudicated} annotated paragraphs,
from 175 news articles and covering an average of 7.86 articles per year  (excerpts shown in Tables~\ref{tab:ideo_example4},~\ref{tab:ideo_example3}, and~\ref{tab:ideo_example2}).
These articles originate from 
5 news outlets related the US Federal Budget from 1947-1975 covering the
center-left, center, center-right spectrum:
Chicago Tribune (CT), Christian Science Monitor (CSM), the New York Times (NYT), Time Magazine (TM), and the Wall Street Journal (WSJ).

With this data, we reveal lexical insights on the language of ideology across the left-right spectrum and across dimensions. We observe that linguistic use even at word level can reveal the ideology behind liberal and conservative policies (Section~\ref{sec:analysis}). Our framework also enables fine-grained, quantitative analysis of polarization, which we demonstrate in Section~\ref{sec:analysis:polarization}. This type of analysis, if scaled up using accurate models for ideology prediction, has the potential to reveal impactful insights into the political context of our society as a whole. 

Finally, we present baselines for the automatic identification of multi-dimensional ideology at the paragraph level (Section~\ref{sec:models}). We show that this is a challenging task with our best baseline yielding an F measure of 0.55; exploring pre-training with existing data in news ideology/bias identification, we found that this task is distinct from, although correlated with, labels automatically derived from news outlets. \emph{We contribute our data and code 
at \url{https://github.com/bernovie/political-polarization}}.

\section{Setup}\label{sec:background}

Many political scientists and political psychologists argue for the use of \emph{at least} a bidimensional ideology for domestic politics that distinguishes between economic and social preferences \cite{carsey2006changing, carmines2012fits, feldman2014understanding}.\footnote{It is important to distinguish between ideology and several other concepts. (1) \textbf{Partisanship} (party identity)~\cite{campbell1980american}: 
a partisan person changes their ideology when their party changes its ideology, whereas an ideological person changes their party when their party changes its ideology. Partisanship is easily conflated with party ID using a unidimensional conceptualization of ideology, but not with a multi-dimensional one. (2) \textbf{Moral foundations}: \citet{haidt2009above} gave an evolutionary explanation of how human morals, values  and traits such as freedom, safety, harm, care, reciprocity, in-group loyalty, authority, equality are formed.  Since, some scholars have used these traits to predict ideology whereas others have attempted to understand what traits unites people with the same ideology. (3) \textbf{Framing}: frames are used in many ways in political science.  They can refer to different ways scholars describe the same information or when scholars talk about different aspects of a single problem~\cite{chong2007framing}.}
We start with these two dimensions while adding a third dimension, ``Foreign'', when the article tackles foreign issues.

Specifically, our \textbf{annotation task} entails examining a news article and annotating each dimension (detailed below) along three  levels---\emph{liberal, conservative, neutral}---for each paragraph. The neutral level for every dimension is reserved for paragraphs related to a specific dimension but either (a) contain both conservative and liberal elements that annotators were unable to ascertain an ideological dimension with confidence, or (b) do not portray any ideology.
We additionally provide an \emph{irrelevant} option if a dimension does not apply to the paragraph. The three dimensions are:

\noindent\emph{\textbf{Social}:}  While the \textit{(1)} \emph{socially conservative} aspect of this dimension is defined as respect for tradition, fear of threat and uncertainty, need for order and structure, concerns for personal and national security, and preference for conformity, its \textit{(2)} \emph{socially liberal} counterpart has been associated with a belief in the separation of church and state, tolerance for uncertainty and change \cite{jost2009political}.

\noindent \emph{\textbf{Economic}:} Similarly, while the \textit{(3)} \emph{economically conservative} aspect of this dimension refers to motivations to achieve social rewards, power, and prestige such as deregulation of the economy, lower taxes and privatization (i.e., being against deficit) spending and advocating for a balanced budget, its \textit{(4)} \emph{economically liberal} counterpart refers to motivation for social justice and equality such as issues related to higher taxes on rich individuals and businesses and more redistribution.

\noindent \emph{\textbf{Foreign}:}  
After piloting the bidimensional approach on 300 articles, we find that using only 2 dimensions conflates two important aspects of ideology related to domestic economy and foreign trade. 
Tariffs, import quotas, and other nontariff-based barriers to trade that are aimed at improving employment and the competitiveness of the US on the international market did not map well onto the bidimensional framework. After consulting several senior political scientists, we adopted a third dimension that dealt with the markets as well as the relations of the US with the rest of the world. 
While the \textit{(5)} \emph{globalist} counterpart of this dimension accounts for free-trade, diplomacy, immigration and treaties such as the non-proliferation of arms, its   \textit{(6)} \emph{interventionist} aspect is nationalist in its support for excise tax on imports to protect American jobs and economic subsidies and anti-immigration.

With the annotated data, we demonstrate quantitative \textbf{measures of polarization} (Section~\ref{sec:analysis:polarization}) and introduce the \textbf{modeling task} (Section~\ref{sec:models}) of automatically identifying the ideology of the policy positions being discussed.
\section{Data collection and annotation}\label{sec:datacollection}

\paragraph{Raw data}
Since polarization is a \emph{process} that needs to be analyzed \emph{over time} \cite{dimaggio1996have}, our annotated articles are sampled from a diachronic corpus of 1,749 news articles across nearly 3 decades (from 1947 till 1974). Articles in this corpus are from political news articles of \citet{desai2019adaptive} from the Corpus of Historical American English (COHA, \citet{davies2012expanding}) covering years 1922-1986. These 1,749 articles are extracted such that: (1) they cover broad and politically relevant topics (ranging from education and health to economy) but still share discussions related to the federal budget to make our annotations tractable\footnote{Because federal budget stories touch on all aspects of the federal activity, this topic appeals to both liberal and conservative media and thus can provide a good testing ground to showcase our proposed ideological annotation method.}; (2) balanced in the number of articles across 5 news outlets with center-left, central, and center-right ideology (c.f.~Section~\ref{sec:analysis:polarization}):
Chicago Tribune (CT), Wall Street Journal (WSJ), Christian Science Monitor (CSM), the New York Times (NYT), and Time Magazine (TM). 
A detailed description of our curation process is in Appendix~\ref{app:data_curation}.

The raw texts were not segmented into paragraphs, thus we used Topic Tiling \cite{riedl2012topictiling} for automatic segmentation. Topic Tiling finds segment boundaries again using LDA and, thus, identifies major subtopic changes within the same article. The segmentation resulted in articles with 1 to 6 paragraphs. The average number of paragraphs per article was 4.

\paragraph{Annotation process}

Our team (including a political science graduate student) developed an annotation protocol for expert annotators using definitions in Section~\ref{sec:background}. The annotation process is independently reviewed by four political science professors from two universities in the US who are not authors of this paper; the research area of two of them is ideology and polarization in the US. We will release our full annotation interface, protocol, and procedure along with the data upon publication.

We sampled on average 7.86 articles per year for annotation, for a total of 721 paragraphs across 175 articles. We divided the annotation task into two batches of 45 and 130 articles, the smaller batch was for training purposes. 
 
In addition to the political science graduate student, we recruited three annotators, all of whom are recent Linguistics graduates in the US. The training sessions consisted of one general meeting (all annotators met) and six different one-on-one meetings (each annotator met with another annotator once). During initial training, the annotators were asked to highlight sentences based on which the annotation was performed. 

\begin{table}[t!]
\centering
\footnotesize
\begin{tabularx}{\linewidth}{>{\hsize=.5\hsize\linewidth=\hsize} X  |       >{\hsize=1.5\hsize\linewidth=\hsize}X}
\toprule
Two \vfill dimensions: socially and \vfill economically \vfill liberal &  \textcolor{blue}{\sethlcolor{lightblue}\hl
{... Secretary of Defense Robert S. [Mc-Namara] threw his full support today behind the Administration's drive against poverty. 
...Mr. Mc-Namara said : ``It is the youth that we can expect to be the most immediate beneficiaries of the war on poverty." He said he was endorsing the ``entire program" both as a citizen and as a member of the Cabinet.}} His endorsement came as his fellow Republicans in Congress continued to hammer away at parts of the Administration's antipoverty program. …  \\
\rowcolor{light-gray}
Two \vfill dimensions: socially  and \vfill economically \vfill conservative & \textcolor{blue}{\sethlcolor{lightblue}\hl{The antipoverty program, the Republicans insisted, would undercut the authority of the Cabinet members by making Sargent Shriver a "poverty czar." ``I don't see how you can lie down and be a doormat for this kind of operation. "}}... \\
\bottomrule
\end{tabularx}
\caption{ Excerpts from article \#723847 in COHA. Because the first paragraph calls for minimizing income inequality, it is socially liberal; and because advocating for such a program call for an budgetary expenditure, it is also has an economic liberal dimension. The second paragraph advocates for the exact opposites of the positions in the first paragraph.  Therefore, it is socially and economically conservative. Sentences most relevant to these labels are highlighted.} 
\label{tab:ideo_example3}
\end{table}

\begin{table}[t!]
\centering
\footnotesize
\begin{tabularx}{\linewidth}{>{\hsize=.4\hsize\linewidth=\hsize} X  |       >{\hsize=1.6\hsize\linewidth=\hsize}X}
\toprule
Zero \vfill dimension &  … ``The committee is holding public hearings on President Eisenhower's Economic Report, which he sent to Congress last week. The Secretary's [Humphrey] appearance before the group provided an opportunity for political exchanges. \\
\rowcolor{light-gray}
One \vfill dimension: economically liberal &  Senators Paul H. Douglas of Illinois, J. W. Fulbright of  Representative Wright Patman of Texas, all Democrats, were active in questioning Mr. Humphrey. \sethlcolor{lightblue}\hl{The Democrats asserted that the Administration's tax reduction program was loaded in favor of business enterprises and shareholders in industry and against the taxpayer in the lowest income brackets}….   \\
One \vfill dimension: economically \vfill neutral  & \sethlcolor{lightblue}\hl{Senator Fulbright .. declare[d] that the problem was to expand consumption rather than production. ... ``Production is the goose that lays the golden egg,`` Mr. Humphrey replied. ``Payrolls make consumers."} \\\bottomrule
\end{tabularx}
\caption{ Excerpts from article \#716033 in COHA. The first paragraph is void of ideology. In the second paragraph the topic is anti tax reduction on businesses, thus it is economically liberal. The third paragraph is simultaneously economically conservative and liberal because one speaker is advocating for decreasing tax on businesses and asserting that production gives an advantage to businesses, the other is advocating for decreasing tax on the poor because they need the income and asserting that healthy businesses are the ones who pay salaries for the low income bracket worker. } 
\label{tab:ideo_example2}
\end{table}

After the annotations of this batch were finalized, the annotators met with the political science student to create ground truth labels in cases of disagreement. 
Then, the three annotators received the second batch 
and each article was annotated by 2 annotators.
This annotation was composed of two stages to account for possible subjectivity.  In stage 1, each annotator worked on a batch that overlapped with only one other annotator.  In stage 2, the two annotators examine paragraphs that they disagree, and met with the third annotator acting as consolidator to adjudicate. Tables~\ref{tab:ideo_example4}. \ref{tab:ideo_example2} and \ref{tab:ideo_example3} 
are examples of adjudicated annotation in the data.

\paragraph{Agreement}
To assess the inter-annotator agreement of stage 1, we report Krippendorf's $\alpha$ \citep{hayes2007answering} for each dimension for the 135 articles after training and before any discussion/adjudication: economic (0.44), foreign (0.68), social (0.39). The agreements among annotators for the economic and foreign dimensions are moderate \& substantial~\cite{artstein2008inter}, respectively; for social, the `fair' agreement was noticed during annotation, and additional discussion for each paragraph was then held. Afterwards, 25 more articles were independently annotated and assessed with an $\alpha$ of 0.53.
Although the agreements were not perfect and reflected a degree of subjectivity in this task, all dimensional labels were \emph{adjudicated} after discussions between annotators.
In total, creating this dataset cost $\sim$150 hours of manual multi-dimensional labeling.

\paragraph{Qualitative analysis of text highlights}

For the 25 articles used in training, all annotators highlighted the sentences that are relevant to each dimension they annotated. This helped annotators to focus on the sentences that drived their decision, and provided insights to the language of ideology, which we discuss here. On average, 21\%--54\% of the sentences in a paragraph were highlighted.

We found entities such as ``President" and "Congress" were the most prevalent in the highlights, and they tackled social and economic issues combined. This is not surprising as it suggests that when the media quotes or discusses the ``President" and ``Congress", they do so with reference to more complex policy issues. In contrast, individual congresspeople tackled mostly economic or social issues.  This also is not surprising as it suggests that individual congresspeople are more concerned with specific issues.  Interestingly,  ``House" and ``Senate" almost always figured more in social issues. This suggest that when news media speaks about a specific chamber, they do so associating this chamber with social issues. Finally, party affiliation was infrequent and was mostly associate with social issues. 
\section{Ideology analyses}\label{sec:analysis}

\begin{table} [t]
\centering
\small
\begin{tabular}{l|l|lll|l}
\toprule
&   \#Docs    & Econ. & Soc. & Fgn. & \textbf{Total}\\
\midrule
CSM &37 & 115 & 63   & 82 &  260 \\
CT     &14 & 48  & 33   & 16  &  97 \\
NYT      &60 & 219 & 114  & 130 & 463 \\
TM       &52 & 134  & 60   & 89  & 283 \\ 
WSJ  & 12 & 42 & 21   & 21  &  84 \\ 
\textbf{Total}      &  175  & 558  & 291  & 338 &  1187\\ 
\bottomrule
\end{tabular}
\caption{Dimensional label counts across all 721 paragraphs in the adjudicated data (there can be multiple dimensions per paragraph).} 
\label{tab:analysis}
\end{table}

\begin{figure}[t!]
    \centering
    \includegraphics[scale=0.39
    ]{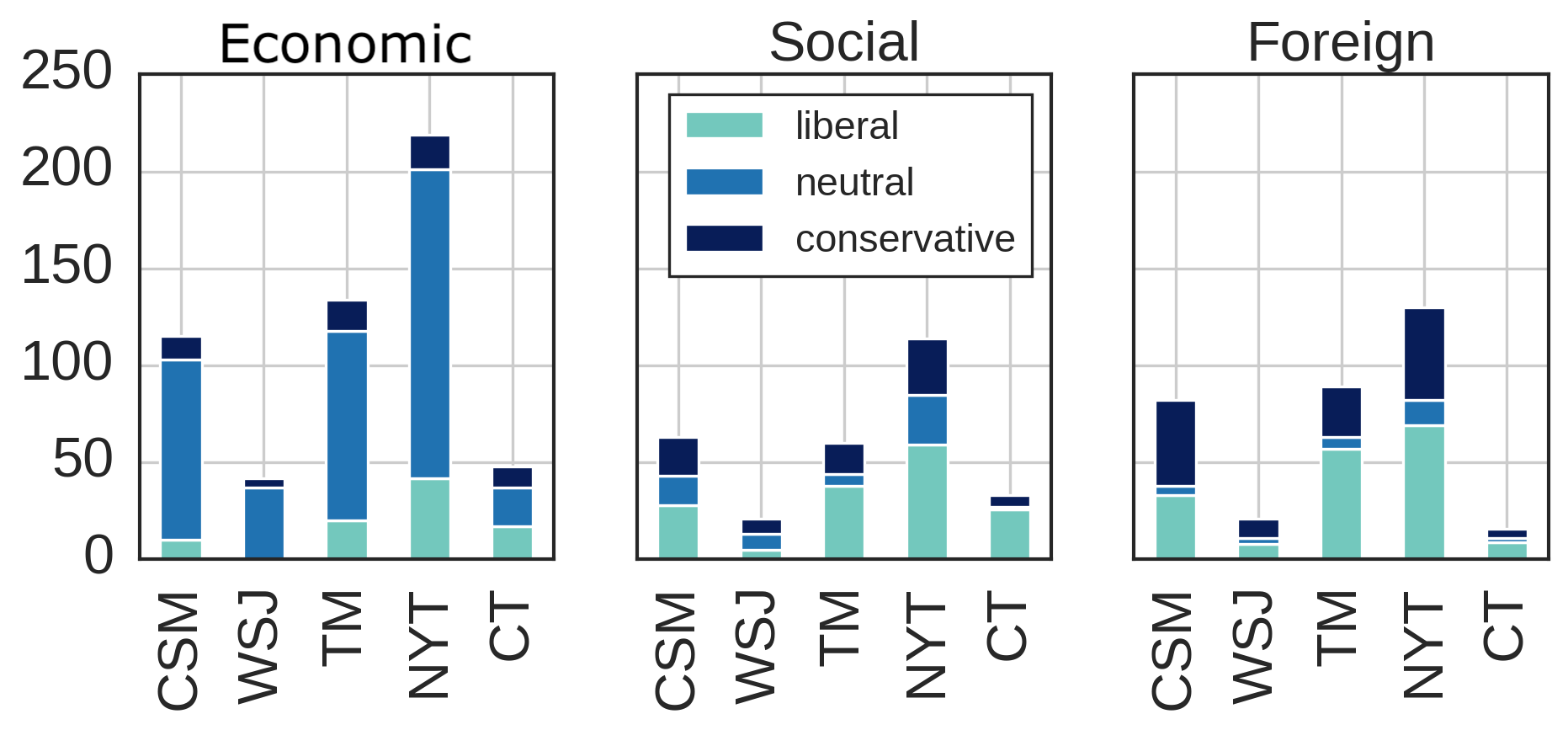}
    \caption{Dimensional label distribution per outlet.}
    \label{fig:label-dist}
\end{figure}

The number of paragraphs per dimension in total is: \textbf{Economics (558), Social (291), Forign (338)}, across the 175 articles. In Table~\ref{tab:analysis} we tabulate this for each of the news outlets.
%
Figure~\ref{fig:label-dist} shows the dimensional label distributions per outlet for each dimension. Expectedly, the dimensional labels often diverge from proclaimed ideology of the news outlet.

\begin{figure}
     \centering
     \begin{subfigure}[b]{0.2\textwidth}
         \centering
         \includegraphics[width=\textwidth]{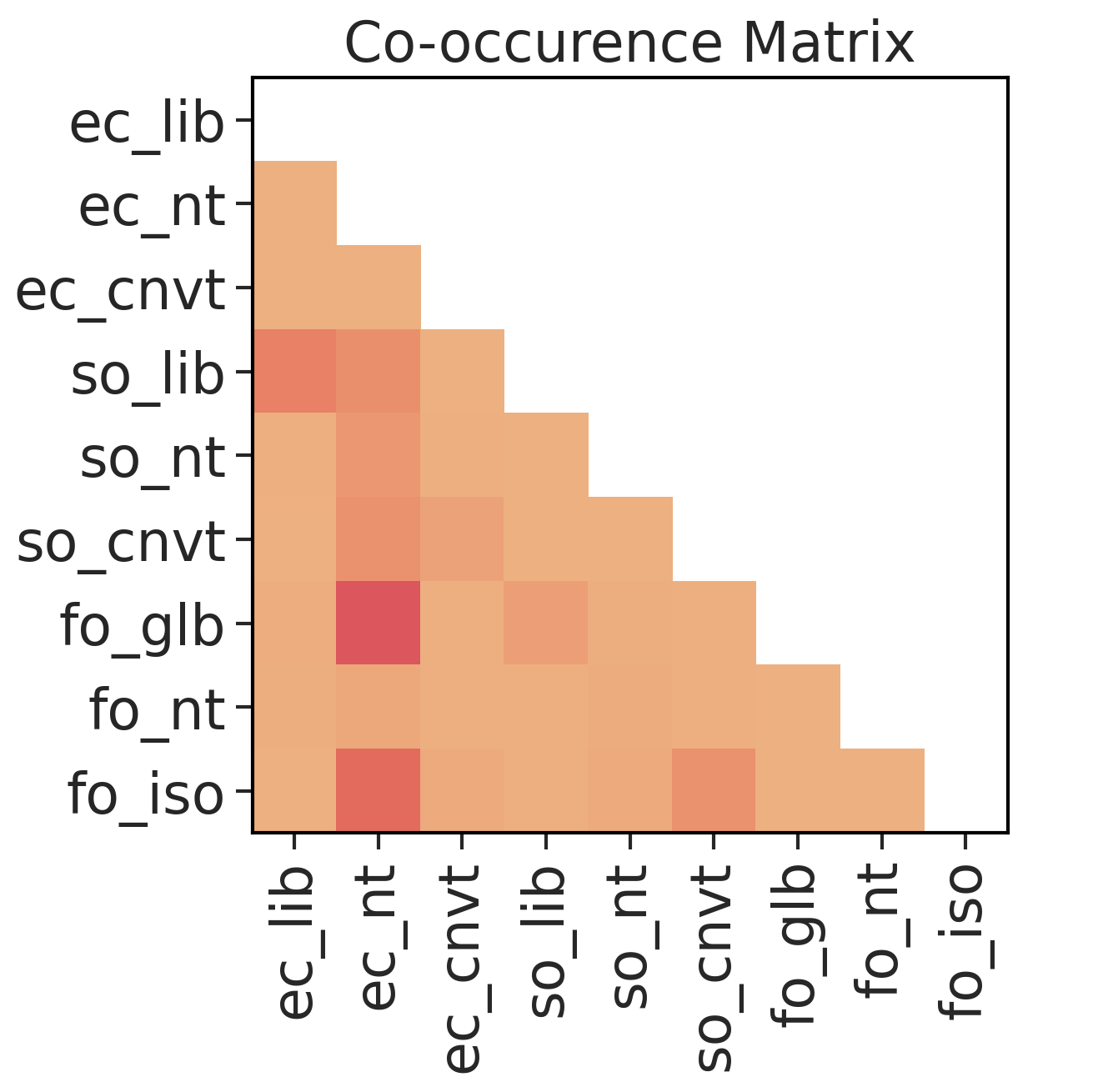}
     \end{subfigure}
     \hfill
     \begin{subfigure}[b]{0.25\textwidth}
         \centering
         \includegraphics[width=\textwidth]{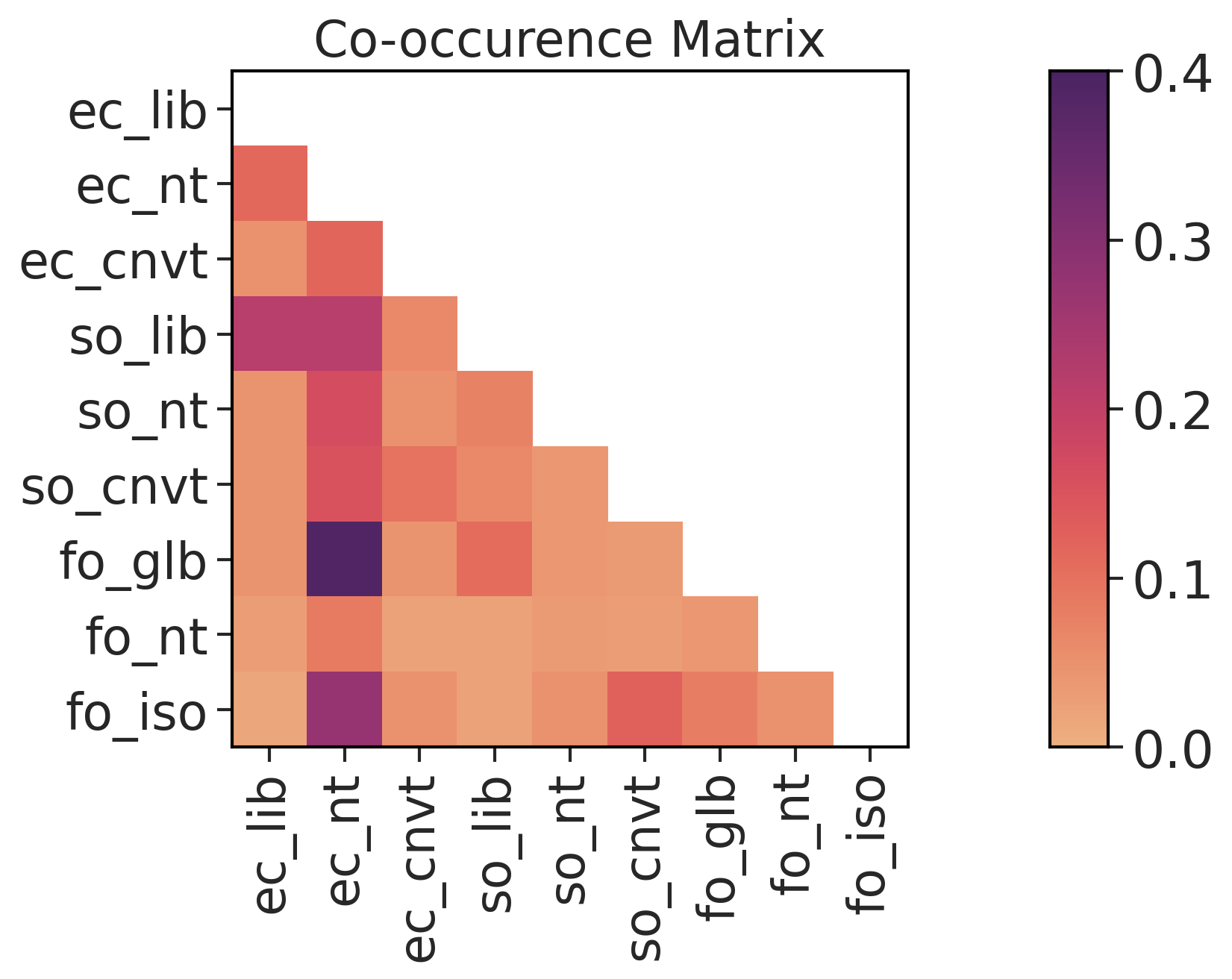}
     \end{subfigure}
     \caption{Co-occurrence matrices on the paragraph level (left) and article level (right)}
     \label{fig:co-occurrance}
\end{figure}

We also analyze the percentage of articles that contain at least one pair of paragraph labels that lean in different directions; for instance, a paragraph with a label of globalist (i.e., liberal) in the foreign dimension and another paragraph with a label of conservative or neutral in the fiscal dimension. The percentage of such articles is 78.3\%. Out of these articles, we examine the average proportion of neutral, liberal, and conservative paragraph labels, and find neutral labels have the highest share (43.27\%), followed by liberal (33.20\%) and conservative (23.53\%). 
In Figure~\ref{fig:co-occurrance} (right), we visualize the percentage of articles where two dimensional labels co-occur within the same article. The figure indicates that ideology varies frequently within an article, 
showing that a single article-level label will not be fine-grained enough to capture variances within an article.

In Figure~\ref{fig:co-occurrance} (left) we also show paragraph-level label co-occurrence. Unlike the article-level, the co-occurrences are less frequent and we are more likely to observe co-occurrences along the same side of ideology. Still, we see interesting nuances; for example, on both the paragraph and the article level, the economic dimension is often neutral, and this tends to co-occur with both liberal and conservative positions in other dimensions.

\paragraph{Lexical analysis}
To understand ways ideology is reflected in text, we also look into the top vocabulary that associates with conservative or liberal ideology. To do so, we train a logistic regression model for each dimension to predict whether a paragraph is labeled conservative or liberal on that dimension, using unigram frequency as features. In Table~\ref{tab:lexical} we show the top most left-leaning (L) or right-leaning (R) vocabulary with their weights. The table intimately reproduces our annotation of ideology.  For example, words like federal and Senator allude to the fact that the topic is at the federal level.  The importance of education and labor to liberals is also evident in the economic and social dimensions in words like school, education, and wage.  The importance of the topic of taxation and defense is evident in conservative ideology in words such as tax, business, missile, and force.

\begin{table}[t]
\centering
\begin{small}
\begin{tabular}{c|p{6.5cm}}
\toprule
\multirow{2}{*}{\rotatebox{90}{Economic}}
& \textbf{C:} mr (5.2), tax (5.0), truman (3.7), business (3.3), billion (3.2) \\
& \textbf{L:} school (-4.3), education (-3.3), commission (-3.2), senator (-3.0), plan (-2.5) \\
\midrule
\multirow{2}{*}{\rotatebox{90}{Social}}
& \textbf{C:} defense (7.5), tax (4.6), air (4.4), billion (3.9), missile (3.8) \\
& \textbf{L:} federal (-3.6), wage (-3.6), would (-3.5), policy (-3.2), labor (-2.9) \\
\midrule
\multirow{2}{*}{\rotatebox{90}{Foreign}}
& \textbf{C:} defense (6.9), force (5.3), north (5.2), air (5.0), vietnam (4.6) \\
& \textbf{L:} aid (-9.3), economic (-5.5), foreign (-5.3), germany (-4.3), make (-4.1)\\
\bottomrule
\end{tabular}
\end{small}
\caption{Words with the most positive and negative weights from a logistic regression model trained to predict liberal/conservative ideology for each dimension.}
\label{tab:lexical}
\end{table}

\section{Polarization}\label{sec:analysis:polarization}
In this section, we demonstrate how our framework can be used to analyze ideological polarization, quantitatively.
To say that two groups are polarized is to say that they are moving to opposite ends of an issue on the multi-dimensional ideological spectrum while, at the same time, their respective political views on ideological issues converge within a group, i.e. socially liberals become also economically liberal \cite{fiorina2008political}.
In political science when ideology is multidimensional, polarization is often quantified by considering three measures that capture complementary aspects~\cite{lelkes2016mass}: (1) \textbf{sorting}~\cite{abramowitz1998ideological} (the extent to which the annotated ideology deviates from an outlet's  proclaimed ideological bias); (2) \textbf{issue constraint}~\cite{baldassarri2008partisans} (a correlational analysis between pairs of ideological dimensions); (3) \textbf{ideological divergence}~\cite{fiorina2008political} (the magnitude of the distance between two groups along a single dimension).
Together these measures describe changes in the ideological environment over time: a \emph{concurrent} increase in 
all three measures indicates polarization in media.

\emph{Limitations:} We use only the fully adjudicated data and refrain from using model predictions, since our baseline experiments in Section~\ref{sec:models} show that predicting ideology is challenging. Hence, the analysis are demonstrations of what our framework \emph{enables}, which we discuss at the end of this section, and the conclusions are drawn for our annotated articles only. We group our data in four-year periods to reduce sparsity.

\paragraph{Measure 1: Sorting} We adapt the sorting principle of \citet{abramowitz1998ideological} to our data and investigate the difference between the proclaimed ideological bias of a news outlet and the ideology of annotated articles from the outlet. To obtain the bias $B_j$ of a news outlet $j$, we average the ratings of each news outlet across common sites that rates media bias (Adfontes, Allsides, and MBFC), yielding: \textbf{CSM (-0.07), CT (0.15), NYT (-0.36), TM (-0.4), WSJ (0.32)} (c.f.~Table~\ref{tab:composite} in  Appendix~\ref{app:composite} for ratings from each site). 

To obtain the overall ideology $I_i^{(j)}$ of article $i$ from outlet $j$, we take the average of liberal (-1), neutral (0), and conservative (1) labels across its paragraphs in all three dimensions. Thus, for each 4-year time period with $m$ articles for outlet $j$, the sorting measure would be the absolute distance of article vs. outlet ideology $|\text{avg}_{i=1}^m(I_i^{(j)})-B_j|/B_j$.

\begin{figure}[t!]
   \centering
   \includegraphics[scale=.25
  ]{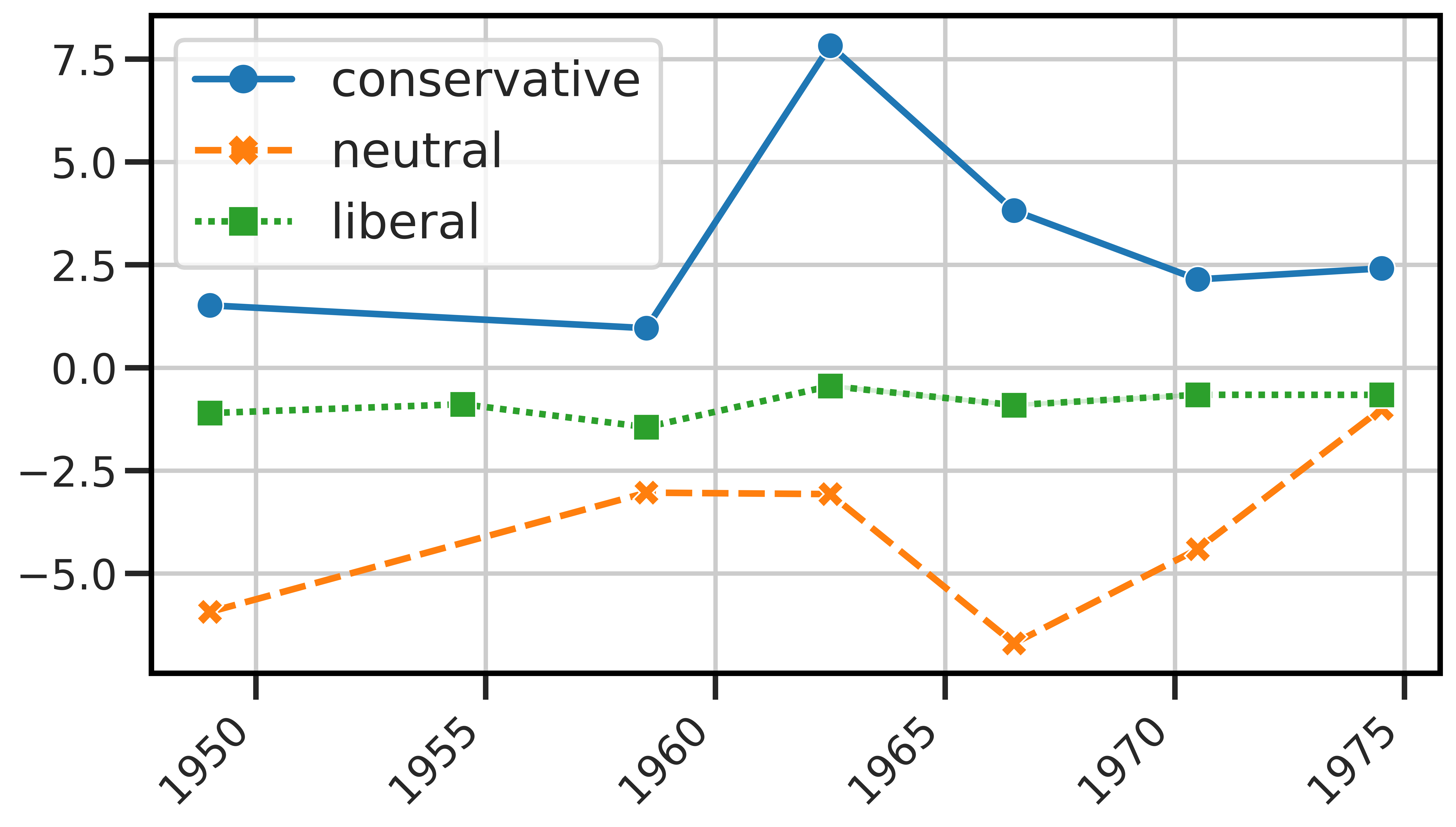}
  \caption{The evolution of the sorting measure, aggregating conservative/neutral/liberal outlets. Moving further away from the zero means articles deviate more from the proclaimed ideology of their outlets.}
   \label{fig:sorting}
\end{figure}

In Figure~\ref{fig:sorting}, we plot the sorting measure, averaged across news outlets of the same proclaimed ideological bias. The figure shows that in our sample of articles,
the left-leaning news outlets were closest to their proclaimed ideological bias measure over time, whereas the neutral outlets were more liberal before 1957 and after 1964.  
The right-leaning outlets were more conservative at that time than their proclaimed ideological bias.

\paragraph{Measure 2: Issue constraint}  This measure refers to the tightness between ideological dimensions over time \cite{baldassarri2008partisans} so as to assess, for example, if socially liberal dimensions are more and more associated with economic liberal dimensions for the news outlets. Concretely, for each article we derive its ideology along a single dimension as the average of paragraph annotations along that dimension. We, then, calculate the Pearson correlation between the article ideology of each pair of dimensions, over all articles from one outlet in the same period.

\begin{figure}[t!]
    \centering
    \includegraphics[scale=2
    ]{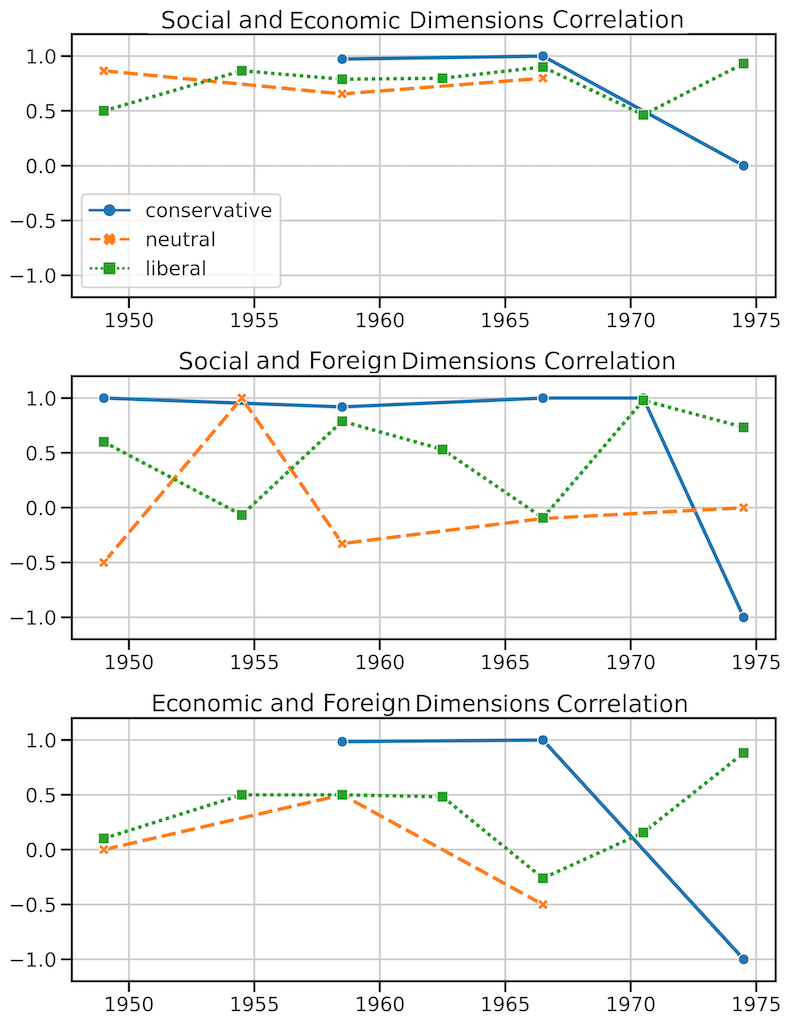} 
    \caption{The evolution of the issue constraint measure, stratified by pairs of dimensions. Higher values mean some dimensions correlate more strongly than others. Due to the lack of articles that simultaneously contains social \& economic dimensions (1st graph), and economic \& foreign dimensions (3rd graph) from conservative outlets their respective blue lines start in 1958.}
    \label{fig:constraint}
\end{figure}

Results in Figure \ref{fig:constraint}, again averaged across news outlets of the same ideological bias, show that for right-leaning media, the correlation between any two dimensions in the annotated data are largely positive (e.g., economically conservative were also socially conservative) until 1967 or 1970. However, for 
proclaimed left-leaning and neutral outlets, the correlations fluctuates especially when considering the foreign dimension.

\paragraph{Measure 3: Ideological divergence}  This measures the distance between two ideological groups on a single dimension~\cite{fiorina2008political}.
We follow~\citet{lelkes2016mass} and calculate the bimodality coefficient~\cite{freeman2013assessing,pfister2013good} per dimension over articles from all news outlets over the same time period. The bimodality coefficient ranges from 0 (unimodal, thus not at all polarized) to 1 (bimodal, thus completely polarized).

\begin{figure}[t!]
    \centering
    \includegraphics[scale=0.25
    ]{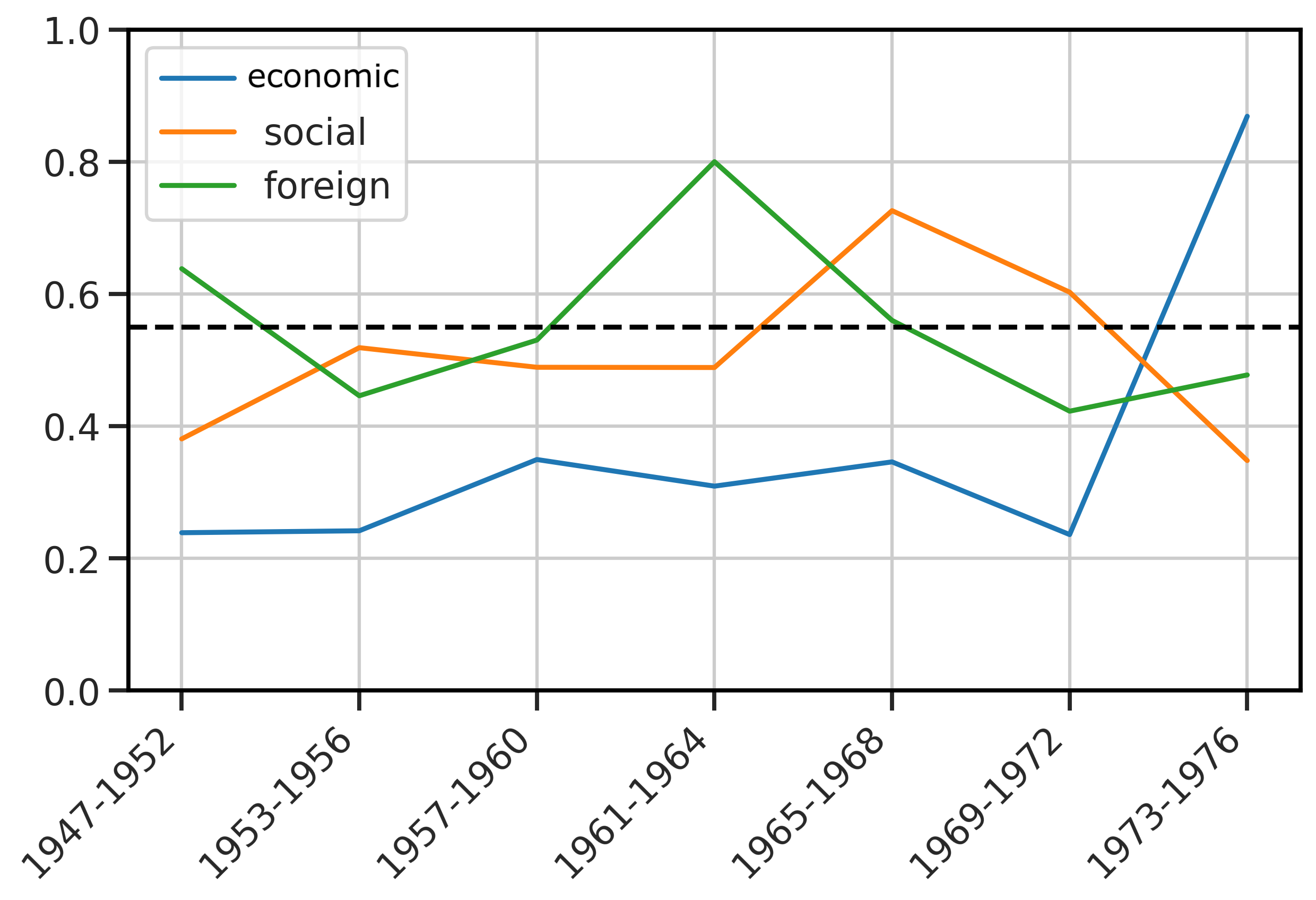}
    \caption{The evolution of the ideological divergence measure stratified by dimension. The dotted line refers to the bimodality threshold \cite{lelkes2016mass}. Higher values mean the ideology of an article along that one dimension is bimodal.}
    \label{fig:divergence}
\end{figure}

Figure \ref{fig:divergence} shows the evolution of the ideological divergence measure of every dimension.  A bimodality measure assesses whether this divergence attained the threshold for the cumulative distribution to be considered bimodal.  Ideological distance, as a result, refers to the three bimodal coefficients. 
We note, for example, that the foreign dimension crossed this threshold between 1956 and 1968. This means that proclaimed left-leaning and right-leaning outlets grew further apart on foreign issues during this time period.

\paragraph{Discussion} 
Taken together, the graphs indicate that the years between 1957 and 1967 are the most noteworthy. During this period, from our sample of articles, we see that polarization was only present in conservative news media because it (1) sorted, as it was significantly more conservative than its composite bias measure, (2) constrained its issues, as evidence by high positive correlation values, and (3) became increasingly bimodal, as the ideological distance between their positions and those of their liberal counterpart on foreign issues increased over time. 
While this conclusion applies to only the set of articles in our dataset, the above analysis illustrates that our framework enables nuanced, quantitative analyses
into polarization. 
We leave for future work, potentially equipped with strong models for ideology prediction, to analyze the data at scale.

\section{Experiments}\label{sec:models}

We present political ideology detection experiments as classification tasks per-dimension on the paragraph level.

We performed an 80/10/10 split to create the train, development, and test sets. The
development and test sets
contain articles uniformly distributed from our time period (1947 to 1974) such that no particular decade is predominant. To ensure the integrity of the modeling task, all paragraphs belonging to the same article are present in a single split. 
The number of examples in the splits for each dimension for the adjudicated data are as follows: for the economic dimension, we had 450 training, 50 development, and 58 test examples. For the social dimension, we had 253 for training, 13 for development, and 25 for testing. For the foreign dimension, we had 266 for training, 33 for development, and 39 for testing.

\subsection{Models}

\paragraph{Recurrent neural networks} We trained a 2-layer bidirectional LSTM \cite{10.1162/neco.1997.9.8.1735}, with sequence length and hidden size of 256, and 100D GloVe embeddings~\cite{pennington-etal-2014-glove}.

\paragraph{Pre-trained language models} We used  BERT-base \cite{devlin2019bert} from  HuggingFace  \cite{wolf2020huggingfaces} and trained
two versions, with and without fine-tuning.
In both cases we used a custom classification head consisting of 2 linear layers with a hidden size of 768 and a ReLU between them. To extract the word embeddings we followed  \citet{devlin2019bert} and used the hidden states from the second to last layer. 
To obtain the embedding of the whole paragraph\footnote{99\% of the paragraphs in the dataset have $\leq$512 tokens.} we averaged the word embeddings and passed this vector to the classification head.

To find the best hyperparameters we performed a grid search in each dimension. For the economic dimension, the best hyperparameters consisted of a learning rate of 2e-6, 6 epochs of training, a gamma value of 2, no freezing of the layers, a 768 hidden size, and 10\% dropout. For the social dimension, the best hyperparameters were a learning rate of 2e-5, 12 epochs, a gamma of 4, no freezing of the layers, a 768 hidden size, and 10\% dropout. Finally, for the foreign dimension the best hyperparameters consisted of a learning rate of 2e-5, 6 epochs, a gamma of 2, no freezing of the layers, a 768 hidden size, and a 10\% dropout.

\paragraph{Focal loss.}
To better address the imbalanced label distribution of this task, we incorporated \emph{focal loss}~\cite{lin2017focal}, originally proposed for dense object detection. Focal loss can be interpreted as a dynamically scaled cross-entropy loss, where the scaling factor is inversely proportional to the confidence on the correct prediction. This dynamic scaling, controlled by  hyperparameter $\gamma$, leads to a higher focus on the examples that have lower confidences on the correct predictions, which in turn leads to better predictions on the minority classes. Since a $\gamma$ of 0 essentially turns a focal loss into a cross entropy loss, it has less potential to hurt performance than to improve it. We found the best $\gamma$ values to be 2 or 4 depending on the dimension.

\paragraph{Task-guided pre-training.} 
We also explored supervised pre-training on two adjacent tasks that can give insights to the relationship between tasks. We used distant supervision that labeled the ideological bias of each article according to that of its news outlet from \url{www.allsides.com}~\cite{kulkarni2018multi}. This procedure allowed us to use the unannotated articles.
\footnote{We also experimented with pre-training on the dataset from \citet{chen2020analyzing}. 
However, because their dataset starts from 2006 (outside of our time domain), this setting performed poorly.}

\begin{table}[t!]
\centering
\footnotesize
\begin{tabular}{l|ccc|c}
\toprule
\multicolumn{1}{l}{} & \multicolumn{1}{l}{\textbf{Econ}} & \multicolumn{1}{l}{\textbf{Social}} & \multicolumn{1}{l}{\textbf{Foreign}} & \multicolumn{1}{l}{\textbf{Average}} \\
\midrule
Majority             & 0.30                       & 0.23                       & 0.25                        & 0.26                        \\
\midrule
BiLSTM               & 0.44                       & 0.37                       & 0.33                        & 0.38                        \\
BERT no-ft           & 0.46                       & 0.31                       & 0.53                        & 0.44                        \\
~~~~+pre-training    & 0.42                       & 0.32                       & 0.46                        & 0.40                        \\
BERT ft              & \textbf{0.64}              & \textbf{0.50}              & \textbf{0.52}               & \textbf{0.55}               \\
~~~~+pre-training    & 0.56                       & 0.47                       & 0.46                        & 0.49                        \\
~~~~-focal loss    & 0.61                       & \textbf{0.50}                       & 0.50                        & 0.54                        \\
\bottomrule
\end{tabular}
\caption{\small Macro F1 of the models averaged across 10 runs.}
\label{tab:results}
\end{table}

\subsection{Results}

Table \ref{tab:results} shows the macro F1 for each configuration, averaged across 10 runs with different random initializations.
The fine-tuned BERT model, with no task-guided pre-training shows the best performance across all 3 ideology dimensions. It is important to note that all the models do better than randomly guessing, and better than predicting the majority class. This shows that the models are capturing some of the complex underlying phenomena in the data. However, the classification tasks still remain challenging for neural models, leaving plenty of room for improvement in future work.

The \emph{BERT ft -focal loss} setting ablates the effect of focal loss against a weighted cross entropy loss.
with weights inversely proportional to the distribution of the classes in the dimension. This loss helped get a bump in the macro F1 score of around 0.1 for each dimension compared to an unweighted cross entropy loss. However, the focal loss gave further improvements for 2 of the 3 dimensions. 
Although task-guided pre-training improved the BERT (no fine-tuning) model for 1 of the 3 dimensions, it led to worse performance than BERT (fine-tuned). The improvement on the no fine-tuning setting indicates that there is a potential correlation to be exploited by the ideology of the news outlet, but such labels are not that informative for multi-dimensional prediction. We hope that this dataset provides a testbed for future work to evaluate more distant supervision data/methods.

\section{Related work}

In contrast to our multi-dimensional approach that examines the ideology of the issue being discussed instead of the author stance, much of the recent work 
in computational linguistics has been dedicated to the latter
(detection of ideological bias in news media) 
while collapsing ideology to one dimension ~\cite{budak2016fair,kulkarni2018multi,kiesel2019semeval,baly2019multi,baly2020we,chen2020analyzing,ganguly2020empirical,stefanov2020predicting}. 
The proposed computational models classify the partiality of media sources without quantifying their ideology~\cite{elejalde2018nature}.

Other researchers interested in the computational analysis of the ideology have employed text data to analyze congressional text data at the legislative level \cite{sim2013measuring, gentzkow2016measuring} and social media text at the electorate level \cite{saez2013social,barbera2015birds}.

In political science, the relationship between (news) media and polarization is also an active area of research. Prior work has studied media ideological bias in terms of coverage 
\cite{george2006new,valentino2009selective}. \citet{prior2013media} argues there is no firm evidence of a direct causal relationship between media and polarization and that this relationship depends on preexisting attitudes and political sophistication. On the other hand, \citet{gentzkow2016measuring} have established that polarization language snippets move from the legislature in the direction of the media whereas \cite{baumgartner1997media} have shown that the media has an impact on agenda settings of legislatures.

\section{Conclusion}
We take the first step in studying multi-dimensional ideology and polarization over time and in news articles relying on the major political science theories and tools of computational linguistics. 
Our work opens up new opportunities and invites researchers to use this corpus to study the spread of propaganda and misinformation in tandem with ideological shifts and polarization. The presented corpus also provides the opportunity for studying ways that social context determines interpretations in text while distinguishing author stance from content.

This work has several limitations. We only focus on news whereas these dynamics might be different in other forms of communication such as social media posts or online conversations, and the legislature. Further, our corpus is relatively small although carefully annotated by experts. Future work may explore semi-supervised models or active learning techniques for annotating and preparing a larger corpus that may be used in diverse applications. 

\section*{Acknowledgements}
This research is partially supported by NSF grants IIS-1850153, IIS-2107524, and Good Systems\footnote{\url{http://goodsystems.utexas.edu}}, a UT Austin Grand Challenge to develop responsible AI technologies.
We thank Cutter Dalton, Kathryn Slusarczyk and Ilana Torres for their help with complex data annotation.
We thank Zachary Elkins, Richard Lau, Beth Leech, and Katherine McCabe for their feedback of this work and its annotation protocol, and Katrin Erk for her comments. Thanks to Pitt Cyber\footnote{\url{ https://www.cyber.pitt.edu}} for supporting this project.
We also acknowledge the Texas Advanced Computing Center (TACC)\footnote{\url{https://www.tacc.utexas.edu}} at UT Austin and The Center for Research Computing at the University of Pittsburgh for providing the computational resources for many of the results within this paper.

\bibliography{acl2020}
\bibliographystyle{acl_natbib}

\clearpage
\appendix

\section{Data curation}\label{app:data_curation}

A diachronic corpus is required to measure and analyze polarization over time \cite{dimaggio1996have}. We 
collect and annotate data across a long period to address the issue of distributional shifts across years \cite{desai2019adaptive,rijhwani-preotiuc-pietro-2020-temporally,bender2021dangers} and help build robust models that can generalize beyond certain periods. 

Additionally, the raw data on top of which we annotate needs to satisfy the following constraints: (1) for human annotation to be tractable, the articles should share some level of topical coherence; (2) for the data to be useful for the larger community, the content should also cover a range of common discussions in politics across the aisle; and (3) the articles should come from a consistent set of news outlets, forming a continuous and ideologically balanced corpus.

We start with the diachronic corpus of political news articles of \citet{desai2019adaptive} which covers years 1922-1986, the longest-spanning dataset to our knowledge. This corpus is a subset of news articles from the Corpus of Historical American English (COHA, \citet{davies2012expanding}).
To extract topically coherent articles, 
we investigate the topics and articles across multiple LDA~\cite{blei2003latent} runs varying the number of topics (15, 20, 30, 50), aiming to arrive at a cluster of topics that share common points of discussion and collectively will yield a sizable number of articles each year from the same news outlets. 

The LDA models consistently showed one prominent topic---the federal budget---across 5 news outlets with balanced ideology (c.f.~Table~\ref{tab:composite}):
Chicago Tribune (CT), Wall Street Journal (WSJ), Christian Science Monitor (CSM), the New York Times (NYT), and Time Magazine (TM). Because federal budget stories touch on all aspects of the federal activity, this topic appeals to both liberal and conservative media and thus can provide a good testing ground to showcase our proposed ideological annotation method. 
In addition to the core federal budget topic (topic 5 of Table~\ref{tab:cluster}), we also include other topics such as health and education that are integral parts of ideological beliefs in the United States, and when discussed at the federal government level, are typically related to the federal budget. 
The top vocabulary of the cluster is shown in Table~\ref{tab:cluster}. 
In an effort to purge articles 
unrelated to the federal budget, we selected only those that contain words such as ``federal'' and ``congress'', and 
excluded those 
that mention state budget, and letters to editors.
(Note that during annotation, we also discard articles that are unrelated to the federal budget.) After this curation, the total number of articles is 5,706 from the 5 outlets.

\begin{table}[t!]
\centering
\footnotesize

\begin{tabularx}{\linewidth}{>{\hsize=.43\hsize\linewidth=\hsize} X  |       >{\hsize=1.55\hsize\linewidth=\hsize}X}
\toprule
Topic1: Trade &    bank, market, farm, loan, export, agricultur, farmer, dollar, food, debt\\ 
\rowcolor{light-gray}
Topic2: \vfill Business &    incom, tax, revenu, profit, corpor, financ, treasuri, pay, sale, bond\\ 
Topic3: \vfill Education &     school, univers, educ, student, colleg, professor, institut, teacher, research, graduat\\ 
\rowcolor{light-gray}
Topic4: \vfill Defense &      nuclear, missil, weapon, atom, test, energi, strateg, bomb, space, pentagon\\
Topic5: \vfill Economy  &      budget, billion, economi, inflat, economic, deficit, unemploy, cut, dollar, rate\\ 
\rowcolor{light-gray}
Topic6: Health/Race &     negro, hospit, medic, health, racial, southern, discrimin, doctor, contra, black\\ 
Topic7: \vfill Industry &    compani, contract, plant, steel, coal, wage, railroad, corpor, manufactur, miner\\ \bottomrule
\end{tabularx}
\caption{ Top words from topics selected in our cluster, from the 50-topic LDA model that yielded the most well-deliminated topics.} 
\label{tab:cluster}
\end{table}

To account for the sparsity of articles in the first decades and their density in later decades, we narrowed down the articles
to the period from 1947 to 1974.  We believe this period is fitting because it includes various ideological combinations of the tripartite composition of the American government, Congress and presidency\footnote{For example, between 1947-49, Congress was Republican and the President was a Democrat while the story flipped between 1955-57.}. The total number of articles in the final corpus of political articles on the federal budget from 1947 to 1974 is 1,749.

\section{Proclaimed ideology of news outlets}\label{app:composite}
\begin{table} [h!]
\centering
\small
\begin{tabularx}{\linewidth}{
>{\hsize=.13\hsize\linewidth=\hsize} X  | 
>{\centering\hsize=.2\hsize\linewidth=\hsize} X|
>{\centering\hsize=.2\hsize\linewidth=\hsize} X |
 >{\centering\hsize=.2\hsize\linewidth=\hsize} X |
>{\hsize=.2\hsize\linewidth=\hsize} X}

 \toprule
      & Adfontes & Allsides & MBFC & \emph{Average}  \\
\midrule
CSM & -.06     & 0.00     & -.16          & -.07 \\
CT           & -.04     & NA     & .34           & .15                \\
NYT            & -.20     & -.5      & -.4           & -.36              \\
TM             & -.10     & -.5      & -.6           & -.4               \\
WSJ       & .15      & .25      & .58           & .32               \\
\bottomrule
\end{tabularx}
\caption{Ideological bias of news outlets from common references of media bias. We use the average in our analyses.} 
\label{tab:composite}
\end{table}

\end{document}